\documentclass[10pt,twocolumn,letterpaper]{article}

\usepackage[pagenumbers]{wacv} 

\usepackage{pifont}        
\usepackage{makecell}      
\usepackage{multirow}      
\usepackage{subcaption}    
\usepackage[table]{xcolor}
\usepackage{booktabs}
\usepackage{array}
\usepackage{tikz}

\usepackage{subcaption}

\definecolor{hmDark}{HTML}{5B1F12}   
\definecolor{hmMid}{HTML}{B9683A}    
\definecolor{hmMidLight}{HTML}{D9A074}
\definecolor{hmLight}{HTML}{ECCDAE}  
\definecolor{hmVLight}{HTML}{F5E4D2} 
\definecolor{hmCream}{HTML}{FAF1E4}  
\definecolor{leakrow}{HTML}{FFE8D6}
\definecolor{lowblue}{RGB}{110,165,220}
\definecolor{highred}{RGB}{255,180,180}
\definecolor{idrow}{RGB}{215,235,255}     
\definecolor{bothrow}{RGB}{225,215,240}   

\newcommand{\cD}[1]{\cellcolor{hmDark}\textcolor{white}{#1}}     
\newcommand{\cM}[1]{\cellcolor{hmMid}#1}                          
\newcommand{\cMl}[1]{\cellcolor{hmMidLight}#1}                    
\newcommand{\cL}[1]{\cellcolor{hmLight}#1}                        
\newcommand{\cVl}[1]{\cellcolor{hmVLight}#1}                      
\newcommand{\cC}[1]{\cellcolor{hmCream}#1}                        
\newcommand{\notrain}[1]{#1$^{\dagger}$}
\newcommand{\trainset}[1]{#1$^{\phantom{\dagger}}$}

\definecolor{wacvblue}{rgb}{0.21,0.49,0.74}
\usepackage[pagebackref,breaklinks,colorlinks,allcolors=wacvblue]{hyperref}

\def\wacvPaperID{55} 
\def\confName{WACV}
\def\confYear{2027}

\title{Not All Similarity Is Leakage:  Distribution-Aware \\ Identity Leakage Assessment in Synthetic Face Datasets}
\author{Paweł Borsukiewicz\\
University of Luxembourg\\
Luxembourg\\
{\tt\small pawel.borsukiewicz@uni.lu}
\and
Daniele Lunghi\\
University of Luxembourg\\
Luxembourg\\
{\tt\small daniele.lunghi@uni.lu}
\and
Wendkûuni C. Ouédraogo\\
University of Luxembourg\\
Luxembourg\\
{\tt\small wendkuuni.ouedraogo@uni.lu}
\and
Fadi Boutros\\
Fraunhofer IGD\\
Darmstadt, Germany\\
{\tt\small fadi.boutros@igd.fraunhofer.de}
\and
Jacques Klein\\
University of Luxembourg\\
Luxembourg\\
{\tt\small jacques.klein@uni.lu}
\and
Tegawendé F. Bissyandé\\
University of Luxembourg\\
Luxembourg\\
{\tt\small tegawende.bissyande@uni.lu}
}

\begin{document}
\maketitle
\begin{abstract}
Synthetic data provides the means for more ethical and less biased facial recognition. Nonetheless, the limitations of generative models and the conditioning process they rely on pose a risk of identity leakage, resulting in synthetic samples that are overly similar to those seen in a generator's training set, mitigating privacy benefits. 
In this work, we show that previously proposed approaches are affected by two key limitations: high classifier dependence and a tendency to identify leakage with respect to real datasets that were not used in training. 
To address these shortcomings, we propose a novel metric called Scaled Overlap (SO), which adapts a non-parametric data-copying test for facial recognition and extends it to the identity level. By leveraging a multi-model setup that compensates for errors in individual classifiers, it provides insights into the true strength of identity leakage.
Our analysis identifies leaking datasets (CemiFace, SFace2, IDNet) with the most acute leakage introduced by generative model conditioning in CemiFace, demonstrates that real datasets do not contribute equally to generator training, and shows that the majority of synthetic images are drawn from multiple identities rather than relying on a single identity.
Ultimately, our study provides new insights into the quality of synthetic datasets and gives more reliable techniques to help the community ensure the trustworthiness of synthetic data.
\end{abstract}  
\section{Introduction}

Facial recognition is a morally ambiguous field. Some view it as a means to ensure safety and order, others scrutinize ethical and privacy aspects~\cite {miethe2025facial} of data collection and processing. 
In a search for privacy-preserving facial recognition solutions~\cite{laishram2025toward} that address concerns of the latter, one stands out: synthetic data. Given its ability to address biases, label noise, and the limitations in the availability of real data~\cite{borsukiewicz2025beyond}, the performance of the models trained on synthetic data has been one of the main factors preventing its use in the early days. 
Now, as fully synthetic models can compete with their real counterparts~\cite{boutros2026idperturb, yeung2024variface}, they are attracting increasing attention~\cite{ButterflyData, wood2021fake, boutros2023synthetic, borsukiewicz2026faces, shahreza2024sdfr, borsukiewicz2026benchmarking}.


Nonetheless, synthetic data is not free from limitations. One critical bottleneck is its reliance on real data for generative model training or conditioning. Namely, a generator may produce a sample that resembles an underlying image, leading to a phenomenon known as identity leakage~\cite{tinsley2021face}. Some of the works have incorporated methods to explicitly address this issue, such as removal of overly similar samples~\cite{wu2025vec2face, sun2024cemiface, papantoniou2024arc2face}, but often with limited success~\cite{kim2024vigface}. 

While previous studies have highlighted the issue~\cite{shahrezaunveiling, borsukiewicz2025beyond}, the analysis of its true extent has been limited, as it primarily focused on generator training and relied on a single pretrained facial recognition (FR) model for assessment. In this work, we propose a more comprehensive approach. 
First, we use five recognizers individually and as an ensemble to find that using previously proposed methods, leakage is observed in datasets unseen by the generator during training, a phenomenon we call \textit{Phantom Leakage}.
By showing its prevalence (100\% leakage at FMR=0.1\% across multiple FR models and dataset pairs), we call into question the faithfulness of the currently established methodology for measuring identity leakage. In search of more trustworthy metrics, we propose Scaled Overlap (SO), accounting for the natural similarity between distinct identities and dataset-specific distributional properties. Contrary to the similarity-based approach, our experimental results demonstrate that SO 
is capable of localizing true instances of leakage.
The main contributions of this work are as follows:
\begin{itemize}
    \item Demonstration of the systematic failure of currently used leakage estimation methods.
    \item Scaled Overlap, an adaptation of the non-parametric data-copying test to face identity space, with an identity-level formulation absent from prior sample-level evaluations.
    \item Comprehensive identity leakage evaluation in 9 synthetic datasets, revealing strong identity-level and limited sample-level leakage in CemiFace, stemming from its generative model conditioning procedure. 
    \item Publicly available replication package\footnote{https://anonymous.4open.science/r/LeakageFinder-E6A3}.
\end{itemize}


Our methodology and findings will help the community evaluate their datasets and, consequently, build trust in synthetic data for facial recognition.
\section{Related Work}

Recent developments in the domain of generative AI, including Generative Adversarial Networks (GANs)~\cite{goodfellow2014generative} and Diffusion Models (DMs)~\cite{ho2020denoising}, have allowed for creation of high-quality synthetic images. This, in turn, has addressed many long-standing shortcomings of real data, including data scarcity and labeling errors. It was also believed to address privacy issues in sensitive domains such as facial recognition. However, recent works~\cite{shahrezaunveiling, tinsley2021face, borsukiewicz2025beyond} have shown that original identities are being revealed through generated samples. In the following subsections, we summarize advancements in the domain of synthetic facial recognition datasets and the identity-leakage threat they pose.

\subsection{Synthetic Facial Recognition Datasets}

Facial recognition practitioners have, for a long time, faced criticism~\cite{smith2022ethical} regarding ethically questionable data acquisition practices, which have typically relied on mass scraping of web databases~\cite{wang2018imdb_face, yi2014casia}.
Synthetic data has emerged as a compelling alternative~\cite{boutros2023synthetic}, mitigating well-known shortcomings of real datasets~\cite{borsukiewicz2025beyond}, including label noise, class imbalance, and potential non-compliance with data protection regulations like the GDPR~\cite{GDPR2016a}.

Early works, such as Syn-1M~\cite{kortylewski2018syn1m} and Syn-Multi-PIE~\cite{colbois2021synmultipie}, have struggled to achieve reliable recognition accuracy, yet subsequent works have continued to push the performance boundaries. More recent methodologies and resulting datasets, including IDPERTURB~\cite{boutros2026idperturb} and VariFace~\cite{yeung2024variface}, have shown remarkable improvements, with the derived models outperforming their counterparts trained on the real CASIA-WebFace~\cite{yi2014casia} dataset. 

While synthetic facial image generation techniques vary widely, there are three main categories of approaches~\cite {borsukiewicz2025beyond}. First focuses on the use of GANs.
SynFace~\cite{qiu2021synface} leveraged DiscoFaceGAN~\cite{deng2020disentangled}, while another set of works (IDNet~\cite{kolf2023identity} and SFace2~\cite{boutros2024sface2}) relied on a StyleGAN2-Ada.
The second group concentrates on the use of conditioned DMs, and it includes datasets such as IDiff-Face~\cite{boutros2023idiff}, DCFace~\cite{kim2023dcface}, CemiFace~\cite{sun2024cemiface}, UIFace~\cite{lin2025uiface}, or ID$^3$~\cite{xu2024text}.
Finally, the third approach used 3D modeling, either independently, as in DigiFace-1M~\cite{bae2023digiface}, or in combination with DMs, as in Digi2Real~\cite{george2025digi2real}, which supplemented it with Arc2Face~\cite{papantoniou2024arc2face}.

With the ability to address the limitations of real data and deliver reliable performance, synthetic data is on track to become the new standard. 

\subsection{Identity Leakage}
\label{sec:id_leakage}


Generative models, both GANs and DMs, have been shown to reproduce samples seen in training~\cite{webster2019detecting, carlini2023extracting}. In our context, we seek samples in which the identities of the generated images are highly similar to those of the real images used to train or condition the corresponding generator, so that facial recognition models assign them sufficiently high similarity scores to be treated as the same identity.
Current protective approaches~\cite{wu2025vec2face, sun2024cemiface, papantoniou2024arc2face} do not fully address this leakage issue, but rather minimize it. 
As such, identities in datasets used to train and/or condition generative models, such as commonly used CASIA-WebFace~\cite{yi2014casia}, FFHQ~\cite{karras2019ffhq}, variants of WebFace~\cite{zhu2022webface260m}, or CelebA~\cite{liu2015faceattributes}, are at risk. 

~\citet{tinsley2021face} have drawn attention to the identity leakage in StyleGAN2~\cite{karras2020analyzing}, trained on FFHQ. Through analysis encompassing 69{,}372 synthetic images and 57{,}094 real samples, they have measured two non-mated distributions: "Real-Real” (R-R) and "Real-Generated” (R-G) across five facial recognition models. Their findings highlighted existing yet limited identity leakage and pointed to disagreements among individual recognizers.


~\citet{singh2024synthprov} have proposed SynthProv, a framework to evaluate identity leakage based on StyleGAN2 latent space, linking synthetic samples with the real identities they follow the most closely.
By combining latent identity dissimilarity and facial recognizer embedding similarity, they computed image-specific provenance scores that indicated which images most influenced each synthetic sample. This approach, while highly interpretable at the individual level, relies on the assumption that the generator is publicly available, which may not hold in practice.

~\citet{shahrezaunveiling} have demonstrated that synthetic facial recognition datasets can expose underlying training identities. In their work, a variant of the Membership Inference Attack (MIA) was designed to identify generated samples across six synthetic facial recognition datasets that are overly similar to the generator's training sets. Using a pretrained ResNet-100~\cite{he2016deep} model, they computed cosine similarity against the training data and identified multiple outliers with unnaturally high scores. 

~\citet{borsukiewicz2025beyond} have used pre-trained IResNet100~\cite{deng2019arcface} to simultaneously evaluate and quantify leakage across all publicly available datasets at the time. The study has found examples of visually similar images in IDNet~\cite{kolf2023identity}, Vec2Face~\cite{wu2025vec2face}, and IDiff-Face~\cite{boutros2023idiff}.  Given that real datasets commonly overlap due to web scraping, the authors have focused only on comparisons involving faces in CASIA-WebFace~\cite{yi2014casia}. 
This simplified approach has omitted potential identity leaks from the source training datasets.




\subsection{Generator Conditioning}



Conditioning is the process of supplying a generator with additional input that steers it toward a specific output rather than a random sample from the learned distribution~\cite{kim2023dcface, boutros2023idiff, lin2025uiface}. It is what allows the generated samples to preserve a consistent identity while other attributes, such as style, vary. 

The origin of this conditioning signal varies widely across the literature~\cite{borsukiewicz2025beyond}. Some works sample identities from the latent space of a pretrained GAN~\cite{colbois2021synmultipie}, or from a disentangled 3DMM identity prior learned from real faces~\cite{qiu2021synface}. Others define identity in the embedding space of a pretrained face recognition model~\cite{boutros2023idiff, sun2024cemiface}, or condition the generator on real identity labels during training~\cite{boutros2024sface2, kolf2023identity}. Still others draw the identity from an unconditional generator and mix it with style references taken from real images~\cite{kim2023dcface, lin2025uiface, boutros2023idiff, xu2024text}, or reuse synthetic identities from prior work, diluting dependence
on real data~\cite{george2025digi2real}.
Arc2Face~\cite{papantoniou2024arc2face} applies Principal Component Analysis (PCA) to the embedding space of a pretrained face recognition model, samples novel identity embeddings within the resulting PCA space, and then uses a pretrained identity-conditioned diffusion model to generate corresponding face images. VIGFace~\cite{kim2024vigface} further advances virtual identity generation in the embedding space by constructing virtual identity prototypes explicitly disentangled from real identities, which are used as conditioning signals for a diffusion-based face generator.

Prior studies on identity leakage (\ref{sec:id_leakage}) have focused primarily on the generator's training data, leaving the conditioning path largely implicit. We argue that conditioning constitutes a second, distinct channel, since the identity signal itself may be derived from, or defined by, real identities. To distinguish the two, we refer to them as \emph{training-data leakage} and \emph{identity-space leakage}, respectively.
\section{Methodology and Experimental Setup}

To ensure a more thorough analysis than the previous works~\cite{borsukiewicz2025beyond, shahrezaunveiling, singh2024synthprov, tinsley2021face},
we design our methodology to support a wide range of evaluation models and synthetic datasets. Through standardized analysis, rooted in distributional properties of real and synthetic data, we aim to demonstrate the true extent of the privacy leakage problem.

\subsection{Classical Leakage Estimation}
\label{sec:Traditional}

To estimate leakage in a classical way, we follow the approaches introduced in prior work~\cite{borsukiewicz2025beyond, shahrezaunveiling}. Our procedure can be briefly summarized as follows:
\begin{itemize}
    \item[\ding{182}] Compute embeddings for real and synthetic datasets.
    \item[\ding{183}] For each model used in evaluation, find key decision thresholds based on real data. 
    \item[\ding{184}] For each synthetic image, find a closest real sample.
    \item[\ding{185}] Compute a proportion of real-synthetic pairs with embedding similarity scores exceeding the decision threshold.
 \end{itemize}

First, for each selected model and dataset (details in~\ref{sec:models} and~\ref{sec:datasets}, respectively), we perform a forward pass to compute embeddings (\ding{182}).
We use non-mated LFW~\cite{huang2008labeled} pairs from its evaluation protocol to determine operational decision thresholds (\ding{183}) based on fixed FMR values (we used Frontex-recommended~\cite{frontex} operational minimum of 0.1\% in the main manuscript; an ablation study of alternative thresholds is provided in the supplementary material).
Even if datasets contain a single image per identity (e.g., FFHQ~\cite{karras2019ffhq}), we can still compute them, since we perform only distinct-identity comparisons. Subsequently, for each synthetic image embedding, we use FAISS~\cite{douze2025faisslibrary} to efficiently find a real image embedding with the highest cosine similarity (\ding{184}). This score is further used to measure leakage as the percentage of synthetic samples for which exists a pair that exceeds previously computed thresholds (\ding{185}). 

\begin{table*}[h]
\centering
\scriptsize
\caption{Synthetic datasets, their deep generative models' training datasets, and the real dataset shaping the identity space (i.e., the pretraining data of the FR encoder used for identity conditioning, or the label source when identities are used directly as classes).}
\label{tab:datasets_used2}
\setlength{\tabcolsep}{3pt}
\begin{tabular}{cc|cc|cc}
\toprule
Synthetic dataset & Total images & Generator training data & Total images & Underlying identity space & Total images \\
\midrule
SynFace~\cite{qiu2021synface}          & 1M     & FFHQ                              & 70k   & FFHQ$^{\dagger}$ & 70k \\
IDiff-Face~\cite{boutros2023idiff}     & 503k   & FFHQ                              & 70k   & MS1MV2$^{\ddagger}$ &  5.8M \\
SynMulti-PIE~\cite{colbois2021synmultipie} & 149k & FFHQ                           & 70k   & FFHQ$^{\dagger}$ & 70k \\
IDNet~\cite{kolf2023identity}          & 1.06M  & CASIA-WebFace                     & 494k  & CASIA-WebFace$^{\star}$  & 494k \\
CemiFace~\cite{sun2024cemiface}        & 550k   & CASIA-WebFace                     & 494k  & WebFace4M$^{\ddagger}$ & 4.2M \\
SFace2~\cite{boutros2024sface2}        & 1.05M  & CASIA-WebFace                     & 494k  & CASIA-WebFace$^{\star}$ & 494k \\
DCFace~\cite{kim2023dcface}            & 550k   & FFHQ + CASIA-WebFace              & 564k  & CASIA-WebFace/WebFace4M$^{\ddagger}$ & 494k/4.2M\\
HyperFace~\cite{shahreza2024hyperface} & 640k   & FFHQ + CelebA + WebFace4M         & 4.5M  & MS1MV2/Glint360K/WebFace42M$^{\ddagger}$ & 5.8M/17.1M/42M \\
Digi2Real~\cite{george2025digi2real}   & 400k   & FFHQ + CelebA + WebFace42M        & 42.2M & Head scans$^{\S}$ & 511 \\
\bottomrule
\end{tabular}
\vspace{2pt}
\raggedright
\scriptsize
\centering
\\
$^{\dagger}$No identity encoder. Identity drawn from a learned prior over real faces.
$^{\ddagger}$Inherited from a pretrained identity encoder. \\
$^{\star}$Identity used as a class label for the identity-conditioned generator. 
$^{\S}$Identities from DigiFace-1M's~\cite{bae2023digiface}. 
\end{table*}

\subsection{Scaled Overlap}
\label{sec:ScaledOvelap}
We propose measuring identity leakage in a synthetic dataset $S$ by examining distributional irregularities relative to a real dataset $R$
in a multi-model setup. 
Our hypothesis is that leakage traces, which can be confused at the level of individual images, should remain consistently detectable at the dataset level with statistical measures.

First, we search for the closest sample in the dataset $R$ for each sample in the datasets $S$ and $R$, excluding mated (same identity) comparisons for real vs. real pairs. To reduce the impact of recognizer-related errors, we leverage an ensemble approach that averages similarity scores across a set of classifiers. The resulting distributions of similarity scores $f_R(x)$ and $f_S(x)$ reflect the natural identity-space diversity of the training dataset and indicate how closely the synthetic data mimics a real-life dataset, respectively. Said resemblance is, to a certain extent, intended, as the generator learns to draw from the real-life distribution it has seen during training. However, when it becomes excessive, we observe identity leakage. 
To quantify this, we adapt the Mann-Whitney U statistic~\cite{mann1947test}, which underlies the data-copying test of Meehan et al.~\cite{meehan2020non}, by replacing their held-out reference sample, unavailable when generators are trained on the full corpus, with leave-one-identity-out nearest-neighbor similarity within the real dataset. We refer to the result as Scaled Overlap (SO):

\begin{equation}
\mathrm{SO}(S, R) = 100 * \int_{-1}^{1} f_S(x)\cdot F_R(x)\,dx
\end{equation}

Using the probability density function $f_S(x)$ of mean similarity scores $x$ observed for a given synthetic dataset and the function $F_R(x)$ that measures the percentile of real data distribution $f_R(x)$ at $x$, we quantify the synthetic distribution in the context of the natural distribution. 
As such, two perfectly overlapping distributions would result in a value of 50, meaning synthetic samples are no more similar to their closest real samples than the closest real samples are to each other. Values above 50 indicate that synthetic samples tend to be more similar to their nearest real counterpart than most similar real samples are to one another, signaling dataset-level leakage. 

We can also apply SO at the identity level by computing identity centroids $c_k$:
\begin{equation}
    \mathbf{c}_k = \frac{1}{|E_k|} \sum_{\mathbf{e}_i \in E_k} \mathbf{e}_i
\end{equation}
where $E_k$ is the set of embeddings belonging to the identity k. To project the newly obtained point on the hypersphere, $c_k$ is subsequently normalized.








\subsection{Models}
\label{sec:models}
To assess the impact of model choice on the leakage analysis, we employ a combination of Convolutional Neural Networks (CNNs)~\cite{krizhevsky2012imagenet} and Vision Transformers (ViTs)~\cite{dosovitskiy2020image}.
We deliberately select a diverse set of pre-trained models with varying capabilities: re-implementation~\cite{facenetpytorch} of FaceNet~\cite{schroff2015facenet}, version trained on CASIA-WebFace~\cite{yi2014casia};
IResNet18~\cite{deng2019arcface, insightface} variant trained on Glint360K~\cite{an2021partial}; SwinFace~\cite{qin2023swinface};
base EdgeFace~\cite{george2024edgeface};
and Glint360K-trained S variant of TransFace~\cite{dan2023transface}. Weights and sources are listed in our replication package. As facial-feature evaluators, these models should be sensitive to genuine instances of leakage, while differences in backbone architecture and training procedures are likely to produce model-specific errors to be partially rectified by an ensemble approach.




\subsection{Datasets}
\label{sec:datasets}
We evaluate our methodology across a wide range of publicly available synthetic datasets (Table~\ref{tab:datasets_used2}) for which generator training data are known. Due to the prevalence of CASIA-WebFace (C-WF)~\cite{yi2014casia}, FFHQ~\cite{karras2019ffhq}, CelebA~\cite{liu2015faceattributes}, and WebFace4M (WF4M)~\cite{zhu2022webface260m} in generator training and their relatively compact size compared with other potential datasets, we primarily focus our analysis on datasets derived from them. Nevertheless, our methodology can be extended to any generative model, provided we have access to its training data and sufficient storage space.

\begin{table*}[ht]
\makebox[\textwidth][c]{%
\centering
\scriptsize
\setlength{\tabcolsep}{2.5pt}
\begin{tabular}{l|cccc|cccc|cccc|cccc}
\toprule
 & \multicolumn{4}{c|}{\textbf{IResNet18}}
 & \multicolumn{4}{c|}{\textbf{TransFace}}
 & \multicolumn{4}{c|}{\textbf{EdgeFace}}
 & \multicolumn{4}{c}{\textbf{5-model ensemble}} \\
\textbf{Dataset}
 & C-WF & CelebA & FFHQ & WF4M
 & C-WF & CelebA & FFHQ & WF4M
 & C-WF & CelebA & FFHQ & WF4M
 & C-WF & CelebA & FFHQ & WF4M \\
\midrule
CemiFace
 & \cD{\trainset{100.0}} & \cC{\notrain{0.4}}  & \cC{\notrain{1.2}}   & \cD{\trainset{100.0}}
 & \cD{\trainset{100.0}} & \cM{\notrain{88.5}} & \cM{\notrain{84.4}}  & \cD{\trainset{100.0}}
 & \cD{\trainset{100.0}} & \cD{\notrain{99.1}} & \cD{\notrain{98.5}}  & \cD{\trainset{100.0}}
 & \cD{\trainset{100.0}} & \cVl{\notrain{19.7}} & \cMl{\notrain{25.1}} & \cD{\trainset{100.0}} \\
DCFace
 & \cD{\trainset{100.0}} & \cC{\notrain{0.2}}  & \cC{\trainset{1.2}}   & \cD{\trainset{100.0}}
 & \cD{\trainset{100.0}} & \cM{\notrain{87.7}} & \cM{\trainset{85.5}}  & \cD{\trainset{100.0}}
 & \cD{\trainset{100.0}} & \cD{\notrain{99.4}} & \cD{\trainset{99.1}}  & \cD{\trainset{100.0}}
 & \cD{\trainset{100.0}} & \cVl{\notrain{13.6}} & \cMl{\trainset{26.2}} & \cD{\trainset{100.0}} \\
Digi2Real
 & \cD{\notrain{100.0}} & \cC{\trainset{0.7}}  & \cC{\trainset{3.6}}   & \cD{\trainset{100.0}}
 & \cD{\notrain{100.0}} & \cM{\trainset{86.4}} & \cM{\trainset{91.2}}  & \cD{\trainset{100.0}}
 & \cD{\notrain{100.0}} & \cD{\trainset{98.3}} & \cD{\trainset{99.4}}  & \cD{\trainset{100.0}}
 & \cD{\notrain{100.0}} & \cVl{\trainset{11.8}} & \cMl{\trainset{33.0}} & \cD{\trainset{100.0}} \\
HyperFace
 & \cD{\notrain{100.0}} & \cC{\trainset{0.3}}  & \cVl{\trainset{7.5}}  & \cD{\trainset{100.0}}
 & \cD{\notrain{100.0}} & \cMl{\trainset{57.4}} & \cM{\trainset{88.1}} & \cD{\trainset{100.0}}
 & \cD{\notrain{100.0}} & \cD{\trainset{97.4}} & \cD{\trainset{99.4}}  & \cD{\trainset{100.0}}
 & \cD{\notrain{100.0}} & \cVl{\trainset{8.8}}  & \cMl{\trainset{42.4}} & \cD{\trainset{100.0}} \\
IDNet
 & \cD{\trainset{100.0}} & \cC{\notrain{0.8}}  & \cC{\notrain{2.0}}   & \cD{\notrain{100.0}}
 & \cD{\trainset{100.0}} & \cM{\notrain{93.7}} & \cM{\notrain{82.5}}  & \cD{\notrain{100.0}}
 & \cD{\trainset{100.0}} & \cD{\notrain{99.1}} & \cD{\notrain{95.9}}  & \cD{\notrain{100.0}}
 & \cD{\trainset{100.0}} & \cMl{\notrain{33.7}} & \cMl{\notrain{23.1}} & \cD{\notrain{100.0}} \\
IDiffFace
 & \cD{\notrain{100.0}} & \cC{\notrain{1.6}}   & \cL{\trainset{19.6}} & \cD{\notrain{100.0}}
 & \cD{\notrain{100.0}} & \cMl{\notrain{51.2}} & \cM{\trainset{89.3}} & \cD{\notrain{100.0}}
 & \cD{\notrain{100.0}} & \cD{\notrain{97.9}}  & \cD{\trainset{99.7}} & \cD{\notrain{100.0}}
 & \cD{\notrain{100.0}} & \cVl{\notrain{18.8}} & \cMl{\trainset{68.4}} & \cD{\notrain{100.0}} \\
SFace2
 & \cD{\trainset{100.0}} & \cC{\notrain{0.7}}  & \cC{\notrain{1.1}}   & \cD{\notrain{100.0}}
 & \cD{\trainset{100.0}} & \cD{\notrain{96.2}} & \cM{\notrain{85.2}}  & \cD{\notrain{100.0}}
 & \cD{\trainset{100.0}} & \cD{\notrain{99.8}} & \cD{\notrain{97.4}}  & \cD{\notrain{100.0}}
 & \cD{\trainset{100.0}} & \cMl{\notrain{37.7}} & \cMl{\notrain{23.2}} & \cD{\notrain{100.0}} \\
SynFace
 & \cD{\notrain{100.0}} & \cC{\notrain{0.5}}  & \cVl{\trainset{9.1}}  & \cD{\notrain{100.0}}
 & \cD{\notrain{100.0}} & \cM{\notrain{87.2}} & \cD{\trainset{98.7}}  & \cD{\notrain{100.0}}
 & \cD{\notrain{100.0}} & \cD{\notrain{99.1}} & \cD{\trainset{100.0}} & \cD{\notrain{100.0}}
 & \cD{\notrain{100.0}} & \cL{\notrain{21.1}}  & \cM{\trainset{76.9}}  & \cD{\notrain{100.0}} \\
SynMulti-PIE
 & \cD{\notrain{100.0}} & \cC{\notrain{0.2}}  & \cVl{\trainset{7.5}}  & \cD{\notrain{100.0}}
 & \cD{\notrain{100.0}} & \cMl{\notrain{55.9}} & \cM{\trainset{90.6}} & \cD{\notrain{100.0}}
 & \cD{\notrain{100.0}} & \cD{\notrain{98.9}} & \cD{\trainset{99.9}}  & \cD{\notrain{100.0}}
 & \cD{\notrain{100.0}} & \cVl{\notrain{15.4}} & \cMl{\trainset{62.7}} & \cD{\notrain{100.0}} \\
\bottomrule
\end{tabular}
}
\centering\noindent\newline\footnotesize{$\dagger$ indicates evaluations where the real dataset was NOT used in either the generator training data or the underlying identity space.}
\caption{Percentage of synthetic samples exceeding FMR=0.1\% threshold across real datasets. Extended table in supplementary material.}
\label{tab:fmr0.1_v2}
\end{table*}
\section{Results}
We designed our study to answer the following research questions:
\begin{itemize}
    \item[\ding{172}] How reliable is the current SOTA leakage analysis?
    \item[\ding{173}] How does Scaled Overlap characterize identity leakage across synthetic face datasets?
    \item[\ding{174}] How does identity leakage manifest in the distributional structure of embeddings?
 \end{itemize}

We start by highlighting the limitations of previous approaches~\cite{borsukiewicz2025beyond, shahrezaunveiling, tinsley2021face} and subsequently demonstrate that SO is a more reliable alternative for leakage estimation. Having established its fidelity, we conduct further statistical tests to analyze leakage across nine synthetic datasets.

\subsection{Limitations of SOTA Leakage Analysis}

\begin{figure}[h]
    \centering
    \includegraphics[width=0.9\columnwidth]{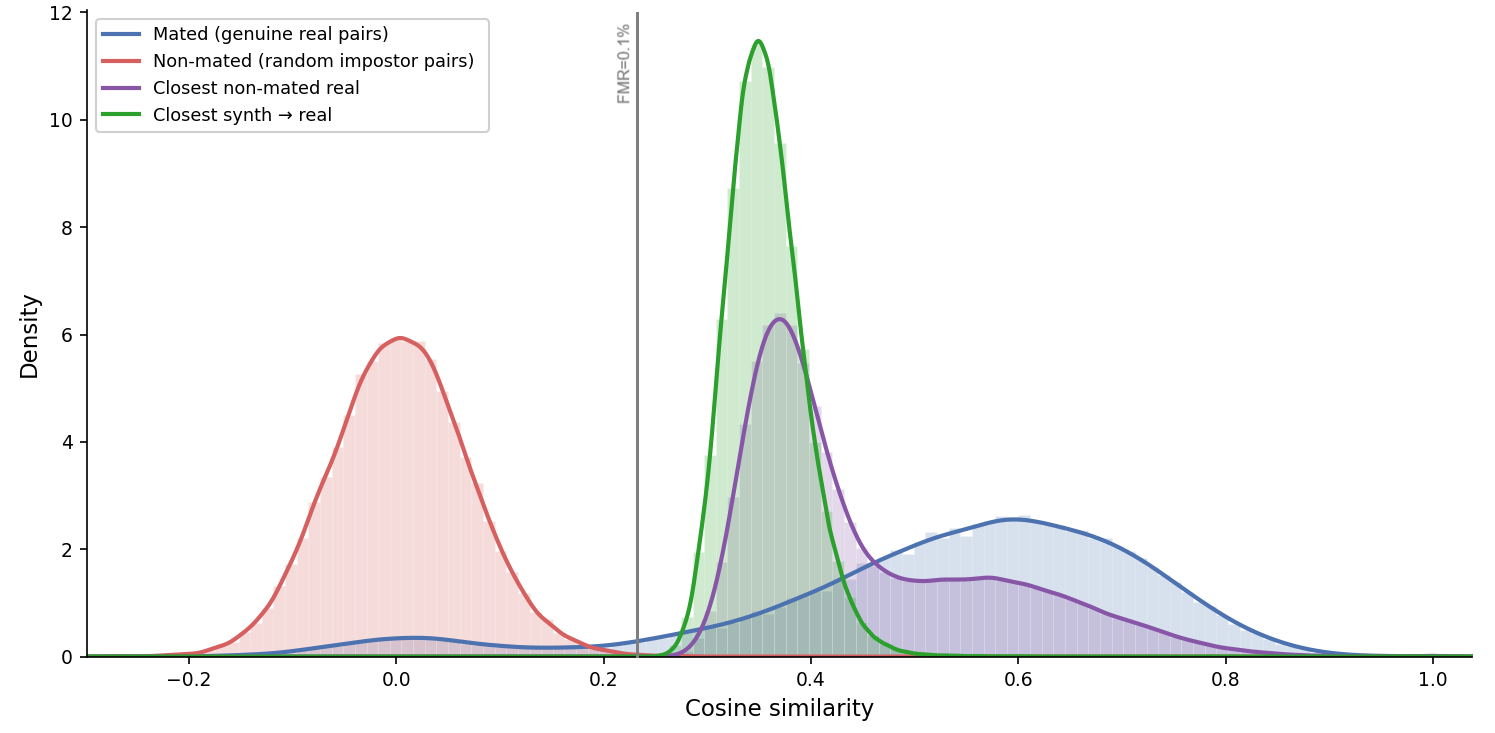} 
    \caption{\textcolor{red}{Mated} and \textcolor{blue}{non-mated} similarity distributions of CASIA-WebFace compared with its \textcolor{violet}{closest non-mated} samples and \textcolor{teal}{closest CemiFace} samples according to EdgeFace. FMR=0.1\% threshold (0.2316) is exceeded by both closest pair distributions.}
    \label{fig:PhantomLeakage}
\end{figure}

We began by computing identity leakage at FMR = 0.1\% as described in~\ref{sec:Traditional}. (Table~\ref{tab:fmr0.1_v2}). A multi-model setup has provided us with very concerning findings.
\noindent\textbf{Leakage values are highly classifier-dependent:} The same dataset, HyperFace, can achieve as low as 0.3 on CelebA, according to IResNet18, and as high as 97.4, according to EdgeFace. This variability demonstrates that a single-model approach with a dependence on a single FMR threshold is fundamentally unreliable and results are meaningless. 
We break down the dependence of leakage on the FMR threshold and model in the supplementary material.
\noindent\textbf{Leakage appears in datasets unseen during generator training and conditioning:} We observe substantial leakage in the datasets where it should not exist, a phenomenon we call \textit{Phantom Leakage}. Most notably, for each evaluated synthetic dataset, all or nearly all (99.9\%+) samples are classified as too similar to C-WF and WF4M.  
Furthermore, the leakage can be deemed extremely high, but the closest non-mated real samples are still more similar than the synthetic samples (Figure~\ref{fig:PhantomLeakage}), putting into question whether observed similarity is a true instance of leakage or simply a naturally occurring phenomenon.

Consequently, we highlight the need for alternative approaches to leakage detection, ones that are free from the limitations of facial recognizer-related inconsistencies and capable of distinguishing real leakage from the naturally occurring similarity between distinct identities.

\subsection{Scaled Overlap}

\begin{figure}[h!]
    \centering
        \includegraphics[width=0.8\linewidth]{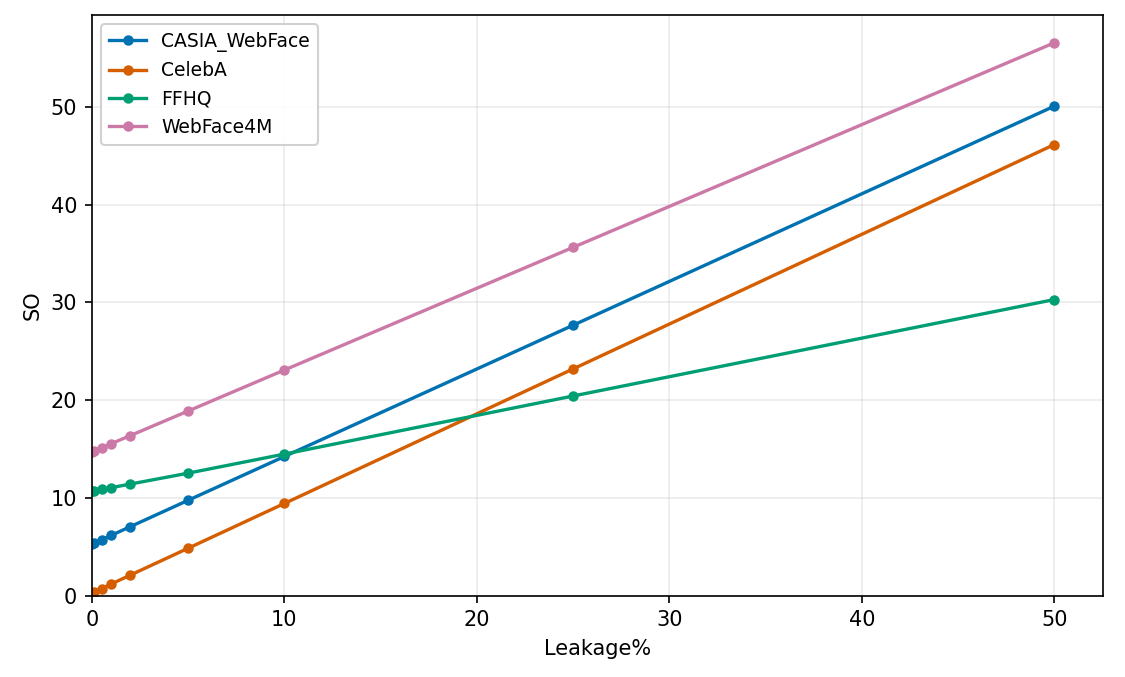}
    \caption{HyperFace with intentionally introduced leakage. SO measured at the image level.}
    \label{fig:blending_v2}
\end{figure}

To showcase the fidelity of the SO metric, we intentionally introduced leakage ($SO_{leak}$) into a synthetic dataset by mixing it with a specified percentage $p$ of real samples from real datasets (Figure~\ref{fig:blending_v2}). These images simulate a scenario in which a generator creates a picture of an identity belonging to its training set. To ensure methodological soundness
, leakage samples are excluded from the original real dataset during evaluation. 
Results for the HyperFace subset (100k samples) show that the SO scales linearly with the percentage of leaking data. 
This is an expected behavior, as:

\begin{table}[t]
\centering
\scriptsize
\setlength{\tabcolsep}{6pt}
\begin{tabular}{llccc}
\toprule
\makecell{\textbf{Synthetic} \\ \textbf{Dataset}} &
\makecell{\textbf{Test} \\ \textbf{Set}} &
\makecell{\textbf{Leakage} \\ \textbf{Candidate}} &
\makecell{\textbf{Sample-level} \\ \textbf{SO}} &
\makecell{\textbf{Identity-level} \\ \textbf{SO}} \\
\midrule
\rowcolor{leakrow} HyperFace    & CelebA &  Yes &  0.2 &  0.0 \\
SynMulti-PIE                    & CelebA &  No  &  0.3 &  0.0 \\
\rowcolor{leakrow} Digi2Real    & CelebA &  Yes &  0.4 &  0.1 \\
DCFace                          & CelebA &  No  &  0.4 &  0.1 \\
SynFace                         & CelebA &  No  &  0.6 &  0.1 \\
CemiFace                        & CelebA &  No  &  0.6 &  0.4 \\
IDiffFace                       & CelebA &  No  &  0.6 &  0.6 \\
IDNet                           & CelebA &  No  &  1.3 &  1.3 \\
SFace2                          & CelebA &  No  &  1.4 &  2.3 \\
HyperFace                       & C-WF   &  No  &  5.3 &  5.1 \\
Digi2Real                       & C-WF   &  No  &  8.2 &  6.8 \\
SFace2                          & FFHQ   &  No  &  4.1 &  9.6 \\
\rowcolor{leakrow} Digi2Real    & FFHQ   &  Yes &  6.8 & 10.5 \\
IDNet                           & FFHQ   &  No  &  4.4 & 10.8 \\
\rowcolor{leakrow} DCFace       & FFHQ   &  Yes &  5.0 & 12.6 \\
CemiFace                        & FFHQ   &  No  &  4.7 & 13.3 \\
\rowcolor{leakrow} HyperFace    & WF4M   &  Yes & 14.7 & 14.2 \\
SynMulti-PIE                    & C-WF   &  No  & 11.8 & 15.2 \\
\rowcolor{leakrow} HyperFace    & FFHQ   &  Yes & 10.4 & 15.5 \\
IDiffFace                       & C-WF   &  No  &  8.4 & 15.6 \\
SynMulti-PIE                    & WF4M   &  No  & 16.9 & 18.1 \\
\rowcolor{leakrow} Digi2Real    & WF4M   &  Yes & 24.3 & 20.4 \\
SynFace                         & C-WF   &  No  & 13.0 & 20.8 \\
\rowcolor{bothrow} SynMulti-PIE & FFHQ   &  Yes & 13.3 & 22.0 \\
\rowcolor{bothrow} SynFace      & FFHQ   &  Yes & 18.9 & 33.3 \\
\rowcolor{leakrow} IDiffFace    & FFHQ   &  Yes & 21.3 & 38.8 \\
SynFace                         & WF4M   &  No  & 30.0 & 39.1 \\
SFace2                          & WF4M   &  No  & 26.9 & 42.4 \\
\rowcolor{bothrow} DCFace       & C-WF   &  Yes & 20.4 & 43.4 \\
\rowcolor{leakrow} CemiFace     & C-WF   &  Yes & 19.3 & 44.4 \\
IDNet                           & WF4M   &  No  & 29.2 & 44.9 \\
\rowcolor{idrow}   DCFace       & WF4M   &  Yes  & 27.5 & 45.3 \\
IDiffFace                       & WF4M   &  No  & 26.8 & 47.6 \\
\rowcolor{bothrow} IDNet        & C-WF   &  Yes & 26.1 & \textbf{57.1} \\
\rowcolor{bothrow} SFace2       & C-WF   &  Yes & 29.7 & \textbf{68.5} \\
\rowcolor{idrow}   CemiFace     & WF4M   &  Yes  & \textbf{53.7} & \textbf{93.3} \\
\bottomrule
\end{tabular}
\caption{sample-level and Identity-level SO. Rows are colored by the relationship between real and synthetic dataset: {\setlength{\fboxsep}{1pt}%
\colorbox{leakrow}{generator training data}, \colorbox{idrow}{identity-space match}, and \colorbox{bothrow}{both}.}}
\label{tab:so_4x6}
\end{table}

\begin{equation}
    SO(p) = (1-p) SO_{synth} + p\cdot SO_{leak}
\end{equation}

At the 0\% mark, where all evaluated samples are synthetic, we observe nonzero overlap scores ($SO_{synth}$), resulting from a combination of the generative model's distribution learning and potential true leakage. 
It is important to note that SO scales differently for FFHQ, with every image treated as a unique identity due to the lack of annotations, confirming that the no-leakage reference point for real data is 50.
Consequently, \textbf{SO should be interpreted as a prevalence-weighted indicator of dataset-level leakage}, rather than a direct count of leaked identities.

Subsequently, we computed SO across all real vs. synthetic dataset pairs to search for leaking datasets (Table~\ref{tab:so_4x6}). Three of the potentially leaking (candidate) dataset comparisons yield the highest SO values, ranging from slightly elevated (IDNet: 57.1, SFace2: 68.5) to highly elevated (CemiFace: 93.3). By surpassing the 50-point mark, synthetic identities are, on average, more similar to the closest real ones than the closest real identities are among themselves. This constitutes \textbf{direct evidence of dataset-level identity leakage}. Additionally, we observed that the strongest instance of leakage (CemiFace) originated from the identity-space match, demonstrating that \textbf{generative model conditioning can be more disruptive than its training}, a factor that was previously overlooked. 

\begin{figure}[t!]
    \centering
    \includegraphics[width=\linewidth]{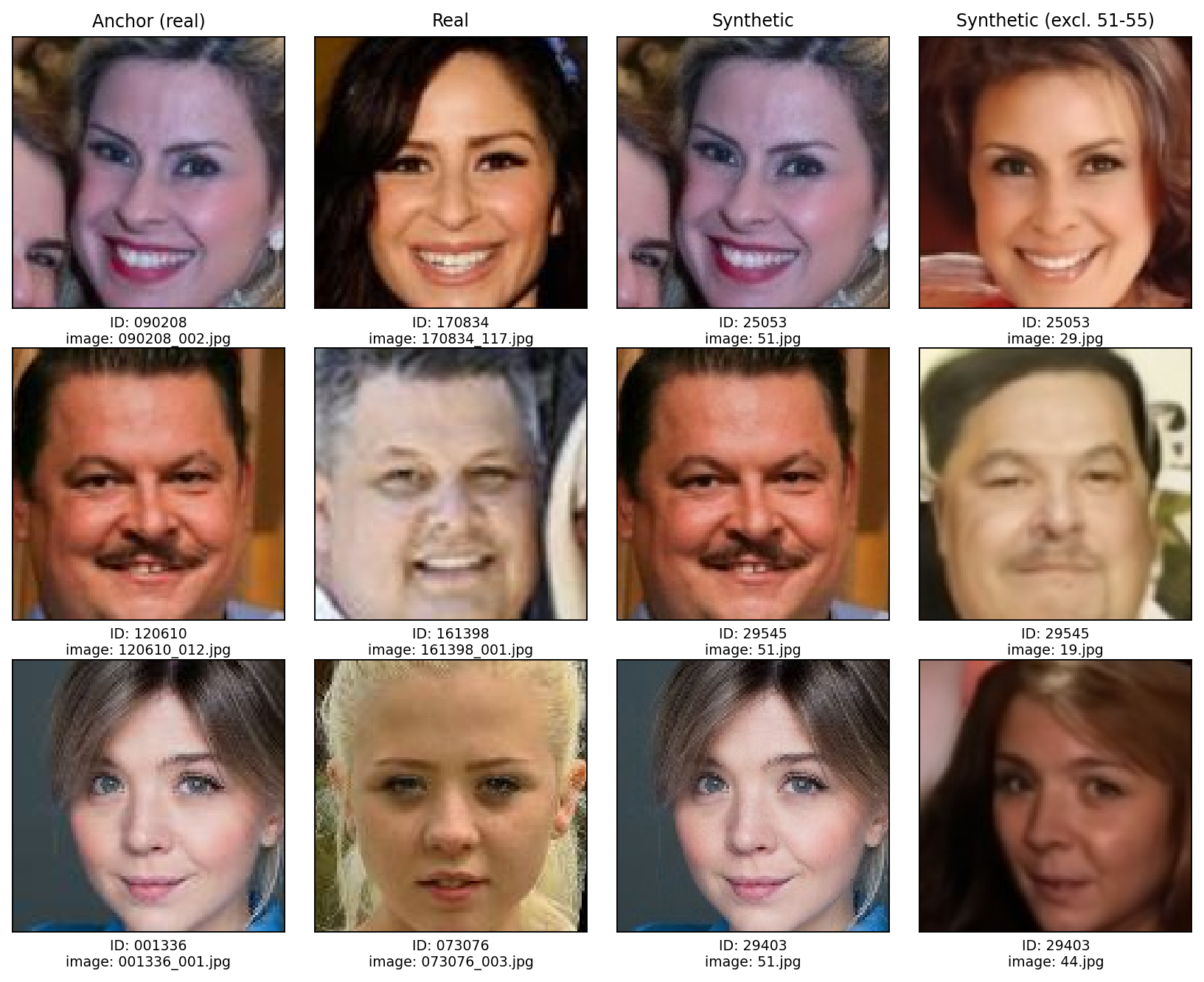}
    \caption{Examples of closest WebFace4M non-mated image vs. closest CemiFace sample (including/excluding samples 51-55) with respect to anchor WebFace4M image.}
    \label{fig:closest_CemiFace}
\end{figure}

Sample-level SO showcases that real pictures are typically more similar to the closest distinct identity image than to a synthetic sample, with an exception of CemiFace scoring 53.7. After analyzing individual closest real and synthetic sample samples for a given real anchor from WebFace4M (Figure~\ref{fig:closest_CemiFace}), we found that \textbf{oversampled images 51-55.jpg across CemiFace identities were copies of real identities seen during conditioning}. While authors~\cite{sun2024cemiface} follow an approach of reusing the same sample five times, previously seen in DCFace~\cite{kim2023dcface}, to shift class center, they diverge as their samples are not generated. This finding validates fidelity of SO in two separate ways: i) identity-level analysis has detected the strong bias towards the underlying real identity amplified by re-use of five copies of real samples; ii) sample-level SO has been able to detect the subtle leakage signal, and since just 5 out of 55 images per synthetic identity are not synthetic, its value exceeds 50 only by a slight margin.
Having repeated SO computation for CemiFace, but excluding these five samples for each identity, we obtained SOs of 0.89 (identity-level) and 0.49 (sample-level), proving that identity leakage does not stem only from the reuse of real copies, but is a dataset-wide issue. 
On the other hand, on an image-by-image basis, synthetic samples almost perfectly follow natural data distribution.
Even though many of the synthesized images beyond samples 51-55 (Figure~\ref{fig:closest_CemiFace}) remain more visually similar to the real anchors than to the closest non-mated real samples, this similarity can be considered natural and is a desired property as long as there is no leakage at the identity level.

Further, even the most outlying datasets:  DCFace (27.5), SFace2 (29.7), and CemiFace (53.7) score significantly lower at the sample level than at the identity level, suggesting that \textbf{leakage in synthetic datasets manifests predominantly at the identity features level rather than at the level of appearance or style}. 
Consequently, even though the majority of individual images remains visually distinct from real photographs, the underlying identity might not.

In datasets, which generators were trained on multiple real datasets, we observe strongly varying sample-level and identity-level SO values: DCFace (C-WF: 20.4 $|$ 43.4 vs. FFHQ: 5.0 $|$ 12.6), HyperFace (CelebA: 0.2 $|$ 0.0 vs. FFHQ: 10.4 $|$ 15.5 vs. WF4M: 14.7 $|$ 14.2) and Digi2Real (CelebA: 0.4 $|$ 0.1 vs. FFHQ: 6.8 $|$ 10.5 vs. WF4M: 24.3 $|$ 20.4). This asymmetry suggests that \textbf{a specific dataset or a subset of real datasets can dominate the generator's learned distribution during training}.

\begin{figure}[t!]
    \centering
    \includegraphics[width=\linewidth]{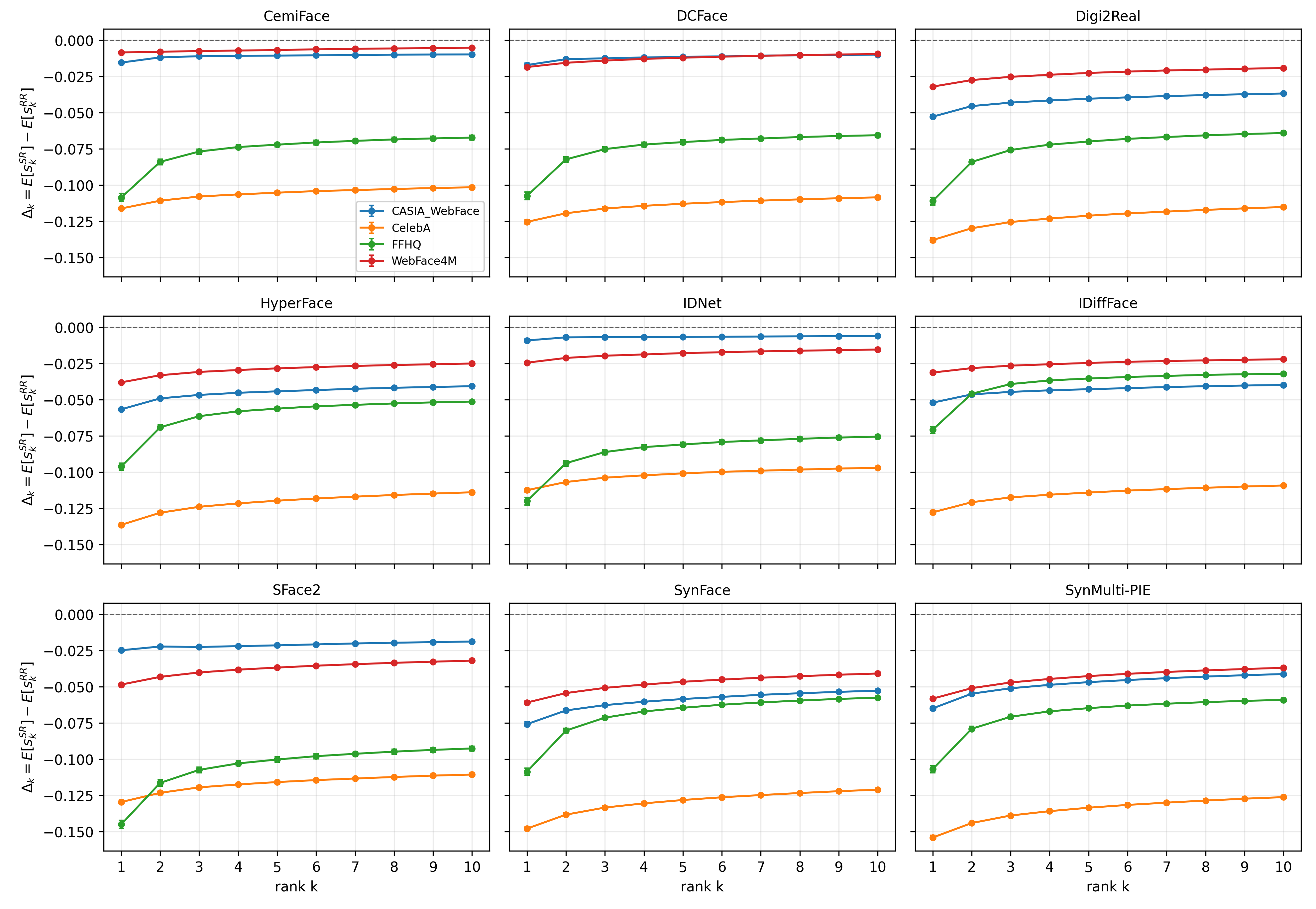}
    \caption{Average cosine similarity difference for top-k most similar real-real and real-synthetic image-level comparisons.}
    \label{fig:sim_topk_v2}
\end{figure}
\begin{figure}[h!]
    \centering
    \includegraphics[width=\linewidth]{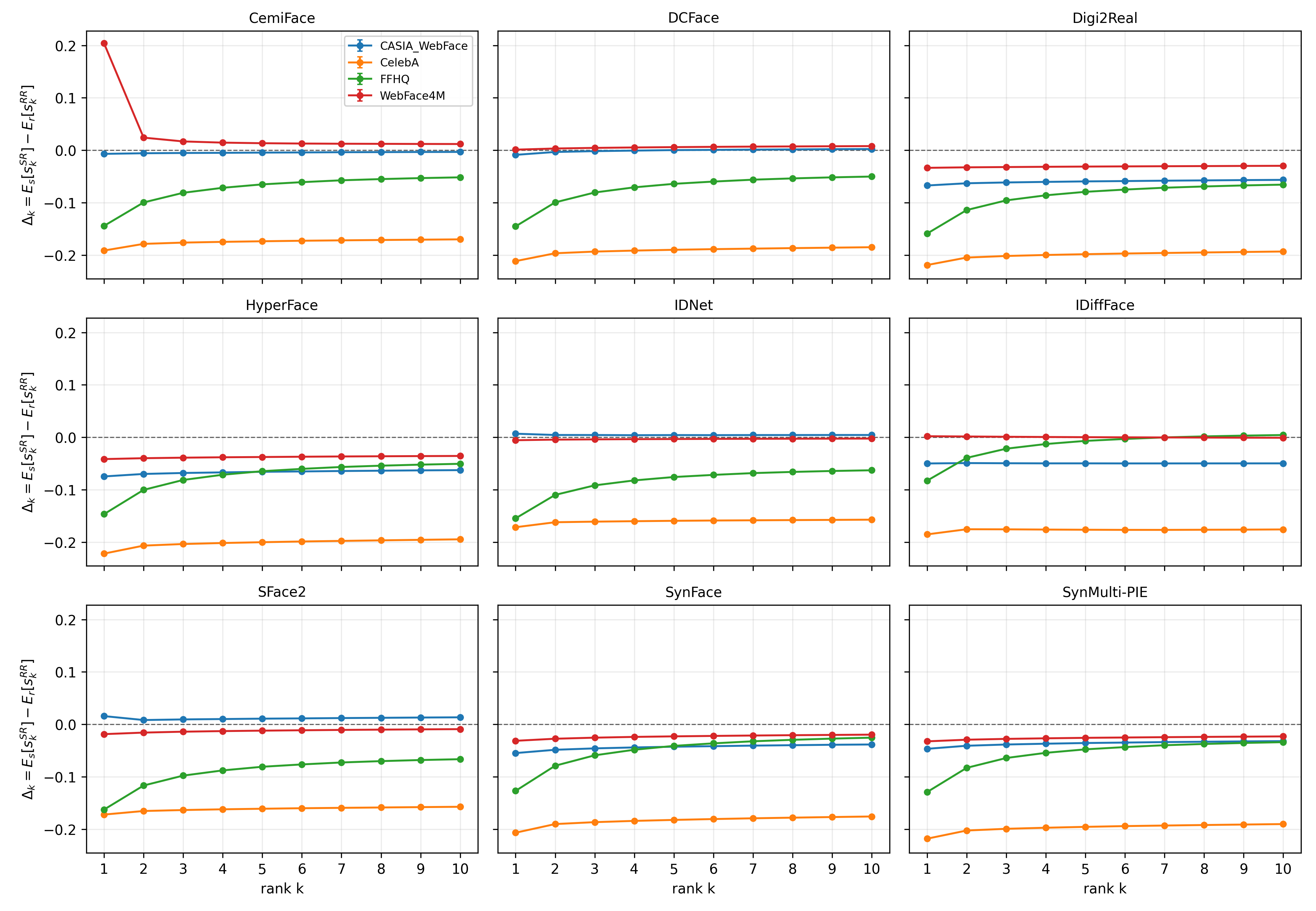}
    \caption{Average cosine similarity difference for top-k most similar real-real and real-synthetic identity-level comparisons.}
    \label{fig:sim_topk_identity_v2}
\end{figure}

\subsection{Distributional Analysis}
\label{sec:4.3}

Another way to probe identity leakage is to examine high-similarity outliers. While prior work~\cite{shahrezaunveiling, borsukiewicz2025beyond} focused on the single closest image pair, we extend this to a neighborhood analysis by tracking, for each real sample, how similarity difference evolves across its $k$-th closest real and $k$-th closest synthetic matches. Figures~\ref{fig:sim_topk_v2} (sample-level) and~\ref{fig:sim_topk_identity_v2} (identity-level) show that the gap $\Delta_k$  between real-to-real and synthetic-to-real similarity typically narrows as $k$ grows, with tight 95\% confidence intervals confirming the stability of this trend. The influence of the training corpus becomes visible further down the ranking: at both levels, FFHQ-trained datasets (HyperFace, IDiffFace, SynFace, SynMulti-PIE, Digi2Real) progressively pull ahead of datasets whose training sets they never saw, while C-WF-trained (CemiFace, SFace2, IDNet) maintain their advantage, indicating that \textbf{training-set exposure influences not only the nearest match but also the clustering strength of synthetic samples surrounding real identities}. 
The leakage signal identified by SO is also reaffirmed at this stage in IDNet and SFace2, which maintain positive $\Delta_k$ across the entire $k \in [1, 10]$ range, meaning their synthetic samples stay closer to real identities than real identities are to each other, even at the tenth neighbor. 
\textbf{The strongest outlier, CemiFace, shows an exceptional bias towards real identities ($\Delta_1$), with an average gap of over 0.2, revealing systematic overfitting to real people.} Conversely, sample-level $\Delta_1$s remain low across all dataset pairs, confirming that leakage manifests in identity space rather than in style.

\begin{figure}[h!]
    \centering
    \includegraphics[width=\linewidth]{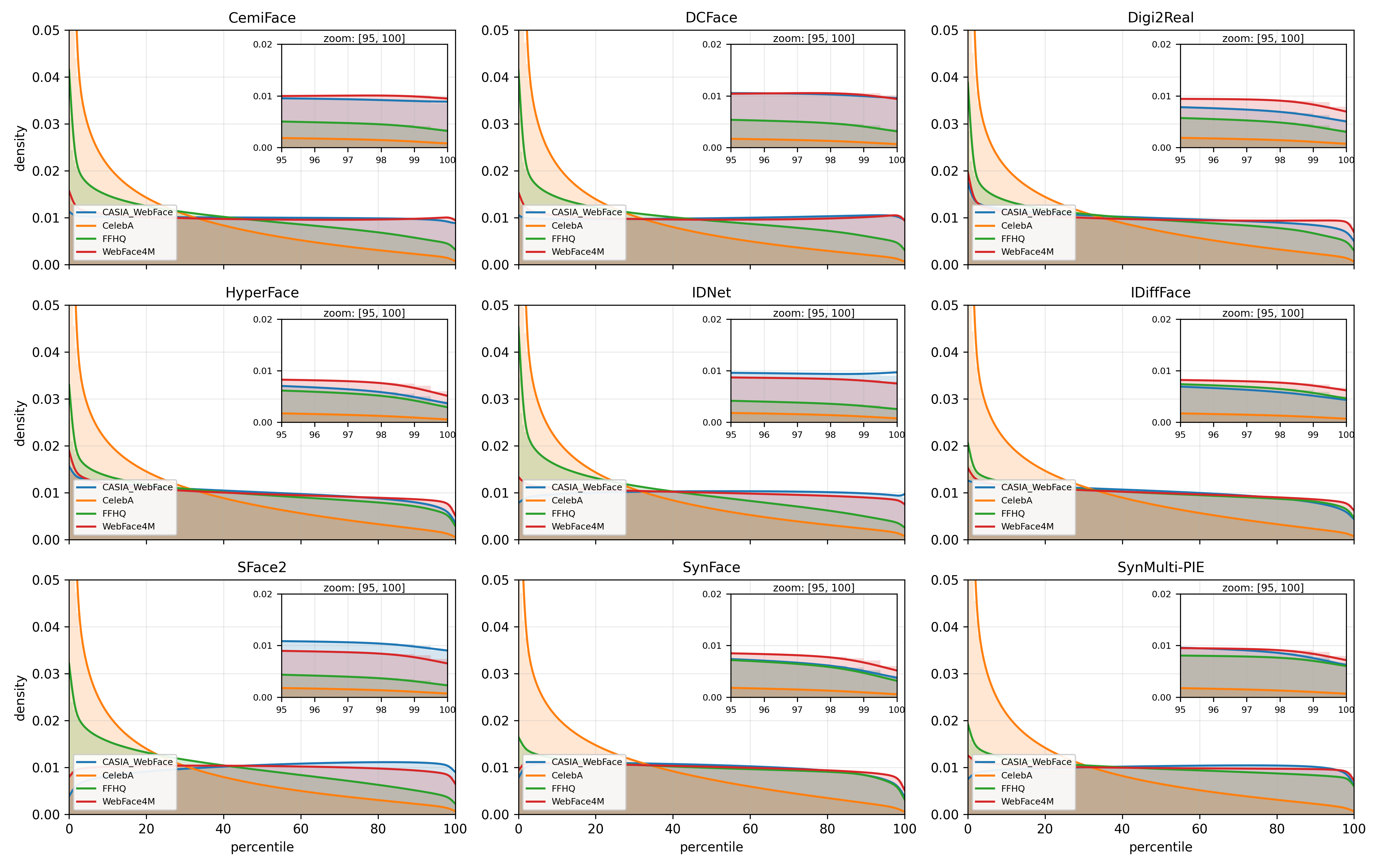}
    \caption{Synthetic-to-real image similarity distribution with real data non-mated self-similarity as a percentile scale.}
    \label{fig:percentile_v2}
\end{figure}

As the $\Delta_k$ analysis captures local clustering, we complement it with a global view of how synthetic embeddings are distributed across the hypersphere. For each real image, we rank all non-mated similarity scores to define a percentile scale, then place each synthetic sample's similarity on that scale to obtain a probability density over percentiles (Figure~\ref{fig:percentile_v2}). Sample-level synthetic data frequently exhibits high density at the low percentiles, indicating a domain gap in a form of structural mismatch with the real distribution. The observed patterns are strongly dependent on the real reference dataset: (i) alignment with CelebA is the weakest across all synthetic datasets, with mass concentrated in the lowest percentiles;
(ii) for FFHQ, a pronounced spike appears near the 0th percentile, but most of the distribution sits around the 0.01 value, with a visible drop-off in the highest percentiles; (iii) synthetic datasets are structurally closest to C-WF and WF4M regardless of generator, with datasets that included C-WF in training (CemiFace, DCFace, SFace2, IDNet) maintaining a steady density up to the top percentiles, while datasets that did not see it drop off noticeably at the tail. Overall, \textbf{synthetic datasets fail to cover the high-similarity tail of real distributions unless the reference was part of their training corpus}. If this stemmed solely from distributional learning, each generator would preserve tails comparably across all three real references. As tail preservation remains selective to the training corpus, it is a sign of memorization rather than generalization, an interpretation consistent with leakage patterns identified by SO and $\Delta_k$.

\begin{figure*}[h!]
    \centering
    \includegraphics[width=\linewidth]{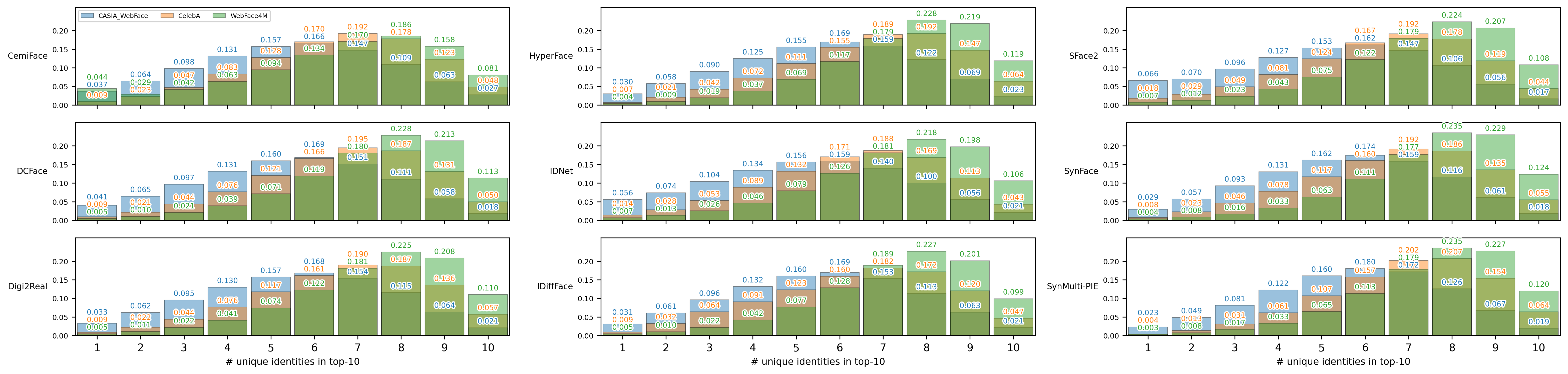}
    \caption{Number of unique identities among top-10 closest real images with respect to synthetic samples.}
    \label{fig:unique_topk_v3}
\end{figure*}

Further, to assess how strongly leakage concentrates on any single person, we measured the number of unique real identities appearing in the top-10 closest matches for each synthetic sample. Since FFHQ is unannotated, it was skipped from this analysis. 
99.68\% (10{,}538 / 10{,}572) of C-WF, 82.23\% (8{,}369 / 10{,}177) of CelebA and 55.53\% (114{,}386 / 205{,}990) of WF4M identities have at least 10 images, ensuring that a focus on a single identity would be detectable. Our results (Figure~\ref{fig:unique_topk_v3}) show that \textbf{synthetic images typically draw from multiple real identities}, with 6-8 distinct identities most commonly represented in the top 10. Fewer than 5\% of synthetic samples are dominated by a single real identity in most datasets. The notable exceptions are SFace2 (6.6\%) and IDNet (5.6\%) against C-WF, consistent with elevated SO and $\Delta_k$ values previously reported for these datasets. 
Even though WebFace4M has a lower fraction of high count identities, it contains approximately 20 times as many identities as CelebA and C-WF, with synthetic images consequently having a much higher number of unique closest real identities. One dominant identity appears at most 0.7\% of the time, except for clearly leaking CemiFace, where this ratio is inflated to 4.4\%.

\section{Discussion}

As our work presents a novel metric, Scaled Overlap, 
we need to address challenges related to differentiating leakage from distribution learning, the closed-source dataset generation procedure, and our dataset-level focus. 

\subsection{Leakage vs. Distribution Learning}

Identity leakage analysis requires clearer guidance on the acceptable level of similarity between real and synthetic datasets. We propose tying it to our dataset-level metric SO. One should consider two complementary directions, identity-level SO and sample-level SO, to prevent overfitting to neither identity-specific facial features nor style-related variables. As such, both sample-level and identity-level similarities should strive to follow the natural data distribution, with respective SOs converging towards 50, yet they should not exceed it.
We argue that this threshold should be slightly lower, 
because: (i) the imperfect distribution learning is a better choice from a privacy perspective than the leakage; (ii) label noise in the real datasets inflates similarity scores for theoretically non-mated real-to-real comparisons, which are in practice mated. 

\subsection{Closed-Source Generator Training}
Recent works in the field of synthetic facial dataset generation~\cite{borsukiewicz2025beyond, korshunov2025investigation, yeung2024variface, sun2024cemiface} have shifted heavily towards diffusion models such as Stable Diffusion~\cite{rombach2022high}.
These, in turn, rely on large amounts of training data that are often not disclosed. Consequently, one would first need to determine whether a given dataset was used during training~\cite{borsukiewicz2026have} before evaluating data leakage using the methodology presented in this paper.

\subsection{Dataset-level Focus}
As SO is designed to measure leakage at the dataset level rather than the individual image level, direct similarity scores comparisons with manual filtering~\cite{shahrezaunveiling} and distributional analysis (as in~\ref{sec:4.3}) remain the most reliable methods to date for identifying specific outliers that exhibit the highest degree of identity leakage. Nonetheless, our analysis, consistent with previous work~\cite{borsukiewicz2025beyond, shahrezaunveiling}, found that classifiers are prone to errors when faces are distorted, occluded, or at extreme angles, leading to substantial variability. Consequently, as proposed in this paper, a multi-judge approach should be leveraged to offset model flaws at the individual sample level as well.




\section{Conclusions}

In this work, we demonstrated two limitations of the current approach to identity leakage estimation in synthetic facial recognition datasets: an overreliance on a single model and a tendency to detect leakage in datasets not used in generator training (\textit{Phantom Leakage}). We addressed both by introducing an evaluation methodology based on an ensemble of five pretrained models and Scaled Overlap, which situates synthetic-to-real similarity against the natural similarity between distinct real identities. Our results show that CemiFace exhibits the strongest dataset-level leakage, with subtler leakage signals observed in SFace2 and IDNet. Leakage itself stemmed primarily from identity-specific features rather than from matching the style of the generator's training or conditioning set. 

Future work should explore two main directions. First, one may leverage distributional data properties presented in this study to design a dataset-wide membership inference attack to determine whether a given real dataset was used for generator training. Second, the leakage with respect to larger training datasets, such as WebFace42M~\cite{zhu2022webface260m}, requires further investigation.

\section*{Acknowledgments}
\begin{itemize}[leftmargin=*]
    \item The authors acknowledge the use of AI-based tools in this work to correct typographical errors as well as to rephrase and refine the clarity of the final manuscript.
    \item This research was supported by the Luxembourg Army.
\end{itemize}

{
    \small
    \bibliographystyle{ieeenat_fullname}
    \bibliography{main}
}

\end{document}